\documentclass[preprint,12pt]{elsarticle}

\usepackage{graphicx}
\usepackage{multirow}
\usepackage{amsmath,amssymb,amsfonts}
\usepackage{amsthm}
\usepackage{mathrsfs}
\usepackage[title]{appendix}
\usepackage{xcolor}
\usepackage{textcomp}
\usepackage{manyfoot}
\usepackage{booktabs}
\usepackage{algorithm}
\usepackage{algorithmicx}
\usepackage{algpseudocode}
\usepackage{listings}
\usepackage{indentfirst}
\usepackage[numbers]{natbib}
\usepackage{float}
\usepackage[section]{placeins}
\usepackage{xurl}
\usepackage{capt-of}
\usepackage{arydshln}
\usepackage{hyperref}
\usepackage{etoolbox}
\usepackage{lipsum}  

\journal{Medical Image Analysis}

\makeatletter
\patchcmd{\pprintMaketitle}
  {\Large\@title\par\vskip18pt}
  {\makebox[\textwidth][c]{\parbox{1.5\textwidth}{\centering\Large\bfseries\@title}}\par\vskip18pt}
  {}{}
\makeatother

\makeatletter
\patchcmd{\pprintMaketitle}
  {\footnotesize\itshape\elsaddress\par\vskip36pt}
  {\footnotesize\itshape\elsaddress\par\vskip18pt
   \par\noindent
   \makebox[\textwidth][c]{%
     \begin{minipage}{0.85\textwidth}
     \centering
     \includegraphics[width=\linewidth]{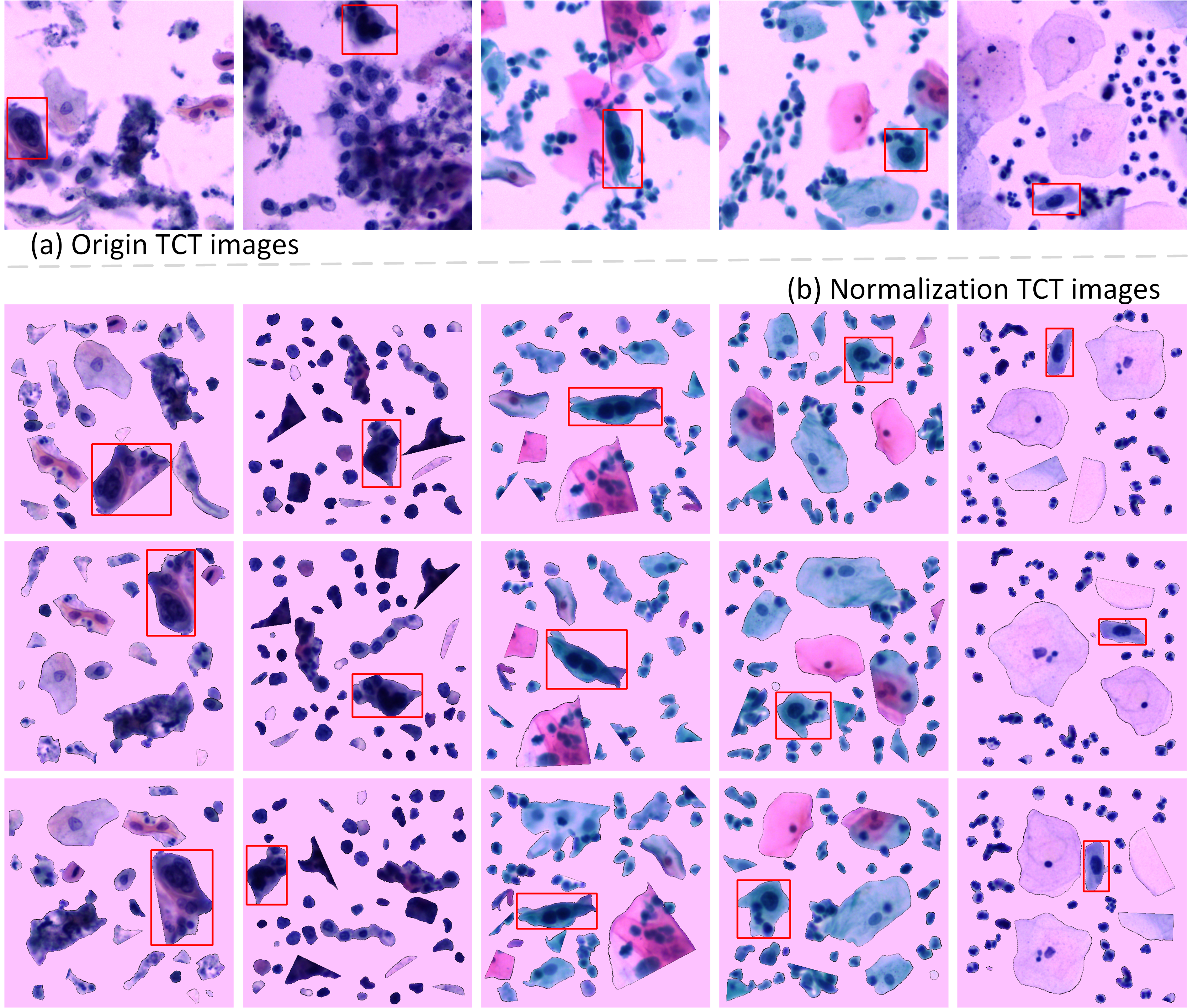}%
     \par\vskip4pt
     \captionof{figure}{Visual demonstration of the proposed Cell-Norm method. (a) The original TCT images and (b) the corresponding Cell-Norm samples.}%
     \label{fig:first}%
     \end{minipage}%
   }%
   \vskip8pt
  }
  {}{}
\makeatother

\makeatletter
\def\ps@pprintTitle{%
     \let\@oddhead\@empty
     \let\@evenhead\@empty
     \def\@oddfoot
       {\hbox to \textwidth%
        {\footnotesize\itshape
         \hfill
         }%
       }%
     \let\@evenfoot\@oddfoot}
\makeatother

\begin{document}

\begin{frontmatter}

\title{C-Norm: Cell-Distribution Normalization Enables\\[3pt]
       Precision Recognition of Medical-Cell Image}

\author[1]{Yang Qianl\fnref{equal}}
\author[2]{Liu Xiany\fnref{equal}}
\author[2]{Dai Daw\corref{cor1}\fnref{equal}}
\ead{dw\_dai@163.com}
\author[2]{Chen Jing\fnref{equal}}
\author[3]{Shen Xiaoj}
\author[1]{Fu Kaiw}
\author[3]{Tang Ming\corref{cor1}}
\ead{tangming@bjmu.edu.cn}
\author[1]{Zou Dongl\corref{cor1}}
\ead{cqzl\_zdl@163.com}

\fnref{equal}
\fntext[equal]{These authors contributed equally to this work.}
\cortext[cor1]{Corresponding author}

\affiliation[1]{organization={Chongqing University Cancer Hospital},
            city={Chongqing},
            country={China}}

\affiliation[2]{organization={Chongqing University of Posts and Telecommunications},
            city={Chongqing},
            country={China}}

\affiliation[3]{organization={Shanghai First Maternity and Infant Hospital, Tongji University},
            city={Shanghai},
            country={China}}

\begin{abstract}
\textbf{Background:} Accurate ThinPrep Cytologic Test (TCT) is pivotal for the early diagnosis of cervical cancer. However, manual cytological examination is labor-intensive and prone to inter-observer variability. Despite the promise of Artificial Intelligence models, the generalization of existing models remains fragile in clinical scenarios. We attribute the limitation of AI models to two fundamental issues: (1) the inherent \textit{spatial distribution imbalance} of cell populations within the TCT images, and (2) the scarcity of large-scale, high-quality datasets due to the requirement for highly specialized expert annotation. These bottlenecks significantly hinder the optimization of large vision models for this specialized task.

\noindent\textbf{Method:} To address these limitations, we propose a \textit{Cell-Distribution Normalization} (C-Norm) method. By decoupling abnormal and normal cells from the original TCT images and re-synthesizing them, this method ensures a uniform distribution of cell populations (See Fig.~\ref{fig:first}), thereby mitigating generalization degradation caused by distribution bias. Building upon this, we integrate the \textbf{YOLOv12} framework with a \textbf{DINOv3} module. This hybrid architecture leverages the advanced detection capability of YOLO models and the superior feature representations of DINOv3 to capture subtle morphological nuances essential for precise recognition of TCT images.

\noindent\textbf{Results:} Extensive experiments demonstrate that our proposed method achieves state-of-the-art performance, significantly outperforming mainstream detection algorithms. These results confirm the effectiveness of our approach, suggesting its potential as a robust assistive tool to alleviate the workload of cytologists and enhance the efficiency of global cervical cancer screening programs. The complete implementation is available in our open-source repository at: \textcolor{blue}{\href{https://github.com/ddw2AIGROUP2CQUPT/Cell-Norm}{https://github.com/ddw2AIGROUP2CQUPT/Cell-Norm}}.
\end{abstract}

\begin{keyword}
Medical Cell \sep Image Classification \sep Deep Learning \sep Data Augmentation \sep Imbalance Learning
\end{keyword}

\end{frontmatter}

\section{Introduction}\label{sec1}

Cervical cancer constitutes a major global health burden, ranking as the fourth most prevalent malignancy in women~\cite{bray2024global,arbyn2023global}. Yet, this disease is largely preventable. Effective screening programs, designed for the early detection of precancerous lesions and nascent malignancies, represent the most powerful tool in its prevention. For the decades, cervical cytology---TCT has served as the principal screening modality. Its implementation has been instrumental in reducing cervical cancer incidence and mortality, underscoring its public health value. Consequently, the precision with which cellular abnormalities are detected and graded in TCT images is fundamental to the success of global efforts to eliminate cervical cancer~\cite{who2022accelerating,wang2026burden,bao2023screen}.

Despite its proven utility, the conventional practice of manual evaluation faces significant limitations. The process is inherently laborious and time-intensive, demanding that highly skilled cytopathologists meticulously examine hundreds of slides each day. \textbf{This manual workflow is not only a substantial drain on healthcare resources but is also prone to significant inter-observer variability, where diagnostic interpretations of the same slide can differ markedly among experts.} Studies have documented kappa values for inter-observer agreement ranging from a low of 0.16 to 0.70, reflecting inconsistent grading of cervical lesions~\cite{viti2025thinlayer,choudhury2024variation}. This subjectivity, compounded by the effects of visual fatigue, contributes to a non-trivial rate of diagnostic errors. False-negative rates, in particular, have been reported to vary widely in different clinical contexts~\cite{macios2022false}, creating the risk of delayed diagnosis and treatment.

Deep learning has yielded impressive results in a variety of medical imaging domains, with models often achieving performance that rival or surpass those of human experts~\cite{liu2025performance,dai2024pallava,tang2024graphconvnet}. However, the advancement of deep learning (such as big models) in this area has been consistently constrained by a primary obstacle: the lack of large-scale and context balanced distribution, publicly accessible datasets, which is insufficient for training the sophisticated, data-intensive models~\cite{zhang2025large,reategui2025review}. Notably, the imbalance in this task refers to the uneven distribution of cell-populations within TCT images (as shown in Fig.~\ref{fig:first}(a)), rather than category-level imbalance across the entire dataset. Conventional data augmentation strategies, which operate at the image or pixel level, fail to mitigate the cell distribution imbalance at the cell-instance level~\cite{shorten2019survey,garcea2023data,stegmuller2024self}. The creation of such datasets is a formidable task, demanding considerable investment of time and specialized knowledge from pathologists to precisely delineate the location and pathological grade of every cellular anomaly.

In this study, we propose a cell populations distribution normalization (Cell-Norm) method to enable cell-instances to be uniformly distributed in TCT images (as shown in Fig.~\ref{fig:first}(b)), which enables the large vision models to achieve superior performance on medical cell detection and recognition tasks. Our method involves segmenting abnormal and normal cells from the original images and recombining them into a new image. And then, we integrate the YOLOv12~\cite{tian2026yolov12} framework with a large vision model DINOv3~\cite{simeoni2025dinov3}, leveraging YOLO's excellent detection capability and DINOv3's powerful feature representation to capture subtle morphological nuances for precise recognition. The contributions of our study can be summarized as:

(1) \textbf{Cell-Norm Method.} \textbf{For medical cell image recognition, we first identify that inconsistent intra-image cell population distribution severely limits the performance of deep learning models.} This fundamental limitation is intractable via conventional data sampling and augmentation strategies. To tackle this issue, we propose the Cell-Norm method, which offers two distinctive advantages. First, it standardizes cell population distribution across all TCT images, \textbf{complying with the essential independent and identically distributed assumption between training and test data that underpins effective machine learning}. Second, it enables cell-level dataset expansion to produce sufficient valid training samples. Collectively, these two unique properties allow our method to resolve inherent dataset defects that cannot be addressed by conventional pixel-level and image-level augmentation pipelines.

(2) \textbf{YoLo-D Model.} \textbf{We integrate the YOLOv12 detection framework with the large vision model DINOv3}, and adopt the proposed Cell-Norm method on the original small-scale dataset. Experimental results demonstrate that our method achieves a significant improvement in abnormal cells detection and localization performance, outperforming all existing state-of-the-art methods.

\section{Related Work}\label{sec2}

\subsection{Deep Learning Models for Medical Cell-Images}\label{subsec:dl}

The application of deep learning to cervical cytology image analysis has evolved substantially over the past decade, progressing from single-cell classification to end-to-end whole slide image (WSI) analysis~\cite{jiang2023systematic,fang2024systematic}. Early deep learning approaches predominantly focused on cell segmentation---isolating individual cell nuclei and cytoplasm from complex backgrounds---as a prerequisite for downstream classification. Encoder-decoder architectures, most notably U-Net~\cite{ronneberger2015unet} and its attention-augmented variants, have been widely adopted for this task due to their ability to capture multi-scale contextual features while preserving spatial resolution, demonstrating strong performance on overlapping and clustered cell configurations~\cite{yin2022u}.

As the field matured, object detection frameworks emerged as a more direct and efficient paradigm for abnormal cell localization, circumventing the dependency on accurate prior segmentation. Two-stage detectors, particularly the Faster R-CNN~\cite{ren2015faster} family with Feature Pyramid Networks (FPN)~\cite{lin2017feature}, have been extensively applied to cervical cytology WSIs. Li et al.~\cite{li2021detection} proposed a Deformable and Global-Context Aware Faster RCNN-FPN (DGCA-RCNN) framework that incorporates deformable convolution layers into the FPN backbone to improve scalability across cell sizes, and introduces a global contextual aware module alongside the Region Proposal Network (RPN) to enhance spatial correlation between foreground and background regions. The DGCA-RCNN achieved a significant improvement over the baseline Faster R-CNN, demonstrating the value of contextual modeling in this domain.

Recognizing that cytopathologists habitually compare suspicious cells against surrounding normal cells to assess abnormality, Liang et al.~\cite{liang2021comparison} introduced the Comparison Detector, which replaces the conventional parametric classifier in Faster R-CNN with a non-parametric comparison-based classifier that evaluates each region proposal against prototype representations of each category. This approach was specifically designed to address the limited data problem in cervical cell detection, achieving significant improvements in mAP and Average Recall (AR) on both small and medium-sized datasets. Building upon this paradigm, Liang et al.~\cite{liang2023exploring} further proposed a cascaded RoI feature enhancement scheme that exploits both cell-level contextual relationships and global image context through two dedicated attention modules---the RoI-Relationship Attention Module (RRAM) and the Global RoI Attention Module (GRAM)---demonstrating superior performance on a large-scale cervical cell detection dataset of 40,000 cytology images.

For WSI-level analysis, Cheng et al.~\cite{cheng2021robust} developed a progressive lesion that combines low- and high-resolution WSI processing with a recurrent neural network (RNN)-based classification model. This work established an important benchmark for clinical-grade automated cervical cancer screening. More recently, Fei et al.~\cite{fei2024distillation} proposed a distillation framework that leverages a Balanced Pre-training Model (BPM)---pre-trained on a generative model-synthesized class-balanced patch dataset---to guide an image-level multi-class lesion cell detector, incorporating a Patch Correlation Consistency (PCC) strategy to exploit inter-cell contextual information. This framework explicitly addresses both incomplete annotations and class imbalance, achieving state-of-the-art performance on multi-class detection.

The emergence of transformer-based architectures~\cite{dosovitskiy2021image} has further enriched the methodological landscape. Qin et al.~\cite{qin2024cell} proposed a cell comparative learning approach for WSI classification that employs pre-trained YOLOX~\cite{ge2021yolox} models to detect both normal and suspected abnormal cells, subsequently leveraging a self-supervised model to extract and compare their features, thereby mimicking the clinical behavior of cytopathologists. At the detection level, Jiang et al.~\cite{jiang2024holistic} introduced a holistic and historical instance comparison framework (HERO) that addresses both cell class ambiguity and class imbalance through a coarse-to-fine comparison strategy, combining RoI-level instance comparison with a class-level memory bank that ensures unbiased sampling of minority class instances. Most recently, Jiang et al.~\cite{jiang2026unicas} presented UniCAS, a cytology foundation model pre-trained on 48,532 cervical WSIs, which supports multi-scale analysis and multi-task inference for cervical cancer screening, candidiasis testing, and cell-level detection across 11 TBS diagnostic categories, establishing a new paradigm for large-scale automated cervical cytology.

\subsection{Class-Level Imbalance and Resampling Strategies}\label{subsec:imbalance}

A fundamental and persistent challenge in the automated medical cell images is the extreme class-level sample imbalance in the clinical data. For example, in a typical TCT slide, normal squamous cells constitute the overwhelming majority, while abnormal cells-particularly high-grade lesions such as HSIL, SCC, and AGC-are exceedingly rare. Among the abnormal categories themselves, the distribution is also highly skewed, with low-grade lesions (ASC-US, LSIL) substantially outnumbering high-grade ones. This long-tailed distribution severely degrades the performance of the current AI models, inducing a strong bias toward majority classes and resulting in high false-negative rates for clinically critical minority categories.

To address the issue of class-level sample imbalance in medical image analysis, three categories of strategies have been explored: cost-sensitive learning (e.g., focal loss~\cite{lin2017focal}, class-weighted cross-entropy), data-level resampling (e.g., oversampling, undersampling, and hybrid methods), and generative data augmentation. Traditional resampling methods such as the Synthetic Minority Over-sampling Technique~\cite{chawla2002smote} and its variants operate in the feature space to synthesize new minority class samples, but their effectiveness is limited when applied to image data with complex semantic content~\cite{eom2023searching}. Deep generative models, particularly Generative Adversarial Networks (GANs)~\cite{goodfellow2014generative}, have emerged as a more powerful alternative for synthesizing realistic minority class-level images. Yu et al.~\cite{yu2021generative} demonstrated that GAN-generated images of abnormal cervical cells can effectively resolve the small sample and class imbalance problem in the cervical cell classification task. Zhao et al.~\cite{zhao2022improving} introduced a dual-transformer framework to generate high-quality, class-balanced cervical cell images and a Tokens-to-Token Vision Transformer for classification tasks.

\subsection{Summary}

Overall, existing work TCT image analysis can be categorized into two primary branches. At the model architecture level, previous studies have continuously advanced detection paradigms; these works mainly focus on designing specialized network modules and comparative learning mechanisms to capture multi-scale cellular features, and relieve feature mismatch caused by overlapping, clustered cells and ambiguous lesion categories. At the data optimization level, existing literature centers on mitigating severe long-tailed class imbalance intrinsic to real clinical TCT specimens. Relevant solutions cover cost-sensitive loss design, feature-space resampling and generative-model-based sample synthesis, aiming to tackle the scarcity of high-grade abnormal cell samples and eliminate model prediction bias toward dominant normal-cell classes. \textbf{While these efforts have advanced TCT image recognition, we argue that they overlook how variations in cell-population distribution affect AI model performance.}

\section{Methods}\label{sec3}

\subsection{Motivation}

\begin{figure*}[h]
    \centering
    \includegraphics[width=\textwidth]{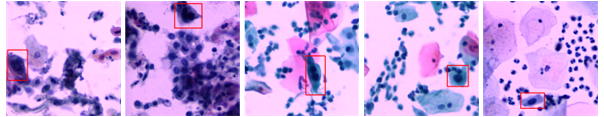}
    \vspace{-3mm}
    \caption{Visual demonstration of severe unevenness in cell population spatial distribution.}
    \label{fig:tct-image}
\end{figure*}

Existing studies on TCT image recognition primarily focus on two directions: refining network architectures to strengthen feature representation, and alleviating dataset class-level imbalance. \textbf{Nevertheless, current advanced approaches overlook inherent the imbalance of cell-population distribution in TCT samples (see Fig.~\ref{fig:tct-image}). These distributional variations persist despite conventional pixel- and image-level augmentation, leading to a mismatch between training and test data that violates the core independent and identically distributed (i.i.d.) assumption. This issue fundamentally constrains the performance of AI models for TCT image analysis.}

Unlike natural images, medical cell images contain only cellular entities. Adjusting the spatial layout of cells will not change their inherent semantic information. To address these long-overlooked yet critical limitations, we present the Cell-Norm method to regularize cell population distribution within individual images (see Fig.~\ref{fig:first}(a) and (b)). For object detection tasks, precise cell boundary segmentation is not a strict requirement, and current medical cell segmentation models have attained sufficient performance to support this task. This approach eliminates the intrinsic heterogeneity of cell distribution and effectively breaks the persistent performance bottlenecks. We believe this method can act as a universal standard preprocessing pipeline for deep learning-based medical cell image analysis.

\subsection{Cell-Norm: Cell-Population Distribution Normalization}\label{subsec:cbs}

\subsubsection{Cell-Instance Segmentation}

\textbf{(1) ROI Extraction:} To exclude background interference and highlight valid content, we first extract regions of interest (ROIs) covering abnormal cells. We then crop local patches around each abnormal cell by expanding outward evenly from all directions on whole-slide images (WSIs), which are used for subsequent analysis. To preserve annotation integrity, we also linearly transform the bounding box coordinates of abnormal cells to match the new ROI coordinate system.

\textbf{(2) Cell-Instance Segmentation:} We adopt the Segment Anything Model (SAM)~\cite{kirillov2023segment} to perform cell-instance segmentation on the cropped ROI images. We first place prompt points uniformly across the ROI image, with the total number adaptively tuned based on image size. For a 640$\times$640 ROI image, the prompts are arranged in a 16$\times$16 grid. SAM model then outputs a separate segmentation mask for each sampled point. We further introduce a three-stage filtering pipeline to discard low-quality masks and boost segmentation accuracy. First, we set the Intersection over Union (IoU) threshold to 0.88 to guarantee strong alignment between predicted masks and real cell contours. Second, a mask stability score of 0.95 is used to ensure reliable segmentation. Third, we limit the minimum mask area to 500 pixels to eliminate fragmented noise and trivial invalid regions. All masks meeting the above criteria are fused via pixel-wise maximum operation into a single binary mask. This mask is overlaid onto the original ROI image: background areas are set to pure black, while the cellular regions retain their original pixel values. 

\subsubsection{Cell-Instance Extraction}

\textbf{(1) Cell Extraction:} Using the segmentation results above, we apply an eight-connectivity domain analysis to locate individual cell regions in the foreground mask. This step distinguishes spatially separate cells and preliminarily resolves overlaps and adjacency between cell bodies. \textbf{This operation enables the identification of spatially independent cell-instance and completes the preliminary separation of overlapping and adjacent cells.} We next conduct morphological filtering to remove residual segmentation noise and further purify isolated cells. Any connected components smaller than 100 pixels are treated as noise and discarded directly. For valid components, we crop individual cell patches using their minimum bounding rectangles. In this process, only pixels within the target component are kept, and all remaining areas are filled with black background. This workflow achieves precise isolation of single cells. Additionally, we record key metadata for each cropped patch, including its position in the original ROI (horizontal and vertical offsets, bounding box width and height) and the actual pixel area of the corresponding cell.

\textbf{(2) Abnormal Cell Matching and Merging:} For training data construction, all cropped single-cell patches are spatially matched with abnormal annotation bounding boxes within the ROI coordinate system to differentiate normal and abnormal cells. We adopt cell coverage rate as the matching metric, defined as the ratio of the overlapping area between the segmented cell bounding box and the annotated abnormal box to the total area of the segmented cell patch. A cell is labeled abnormal when its coverage rate is no less than 0.5; otherwise, it is regarded as a normal cell. Unlike the standard IoU metric that uses the union area as the denominator, our strategy takes the cell patch area for calculation. This tailored design effectively prevents missed abnormal cell detection stemming from oversized annotation boxes, where individual cells only occupy a small fraction of the annotated region and thus yield unreasonably low IoU values.

In practice, one annotated abnormal bounding box may cover multiple segmented cells, especially when cell clusters are split into individual instances by SAM. For this scenario, we merge these discrete cells into a single patch. We first create a blank canvas spanning the combined minimum bounding rectangle of all target cells, then map each cell onto this canvas while preserving its original position within the ROI. This retains their relative layout and yields a complete patch for clustered abnormal cells. After matching, filtering and merging, we obtain two standardized datasets for abnormal and normal cells respectively. All entries are sorted in descending order by patch area, forming well-structured data for subsequent analysis.

\begin{figure*}[h]
    \centering
    \includegraphics[width=\textwidth]{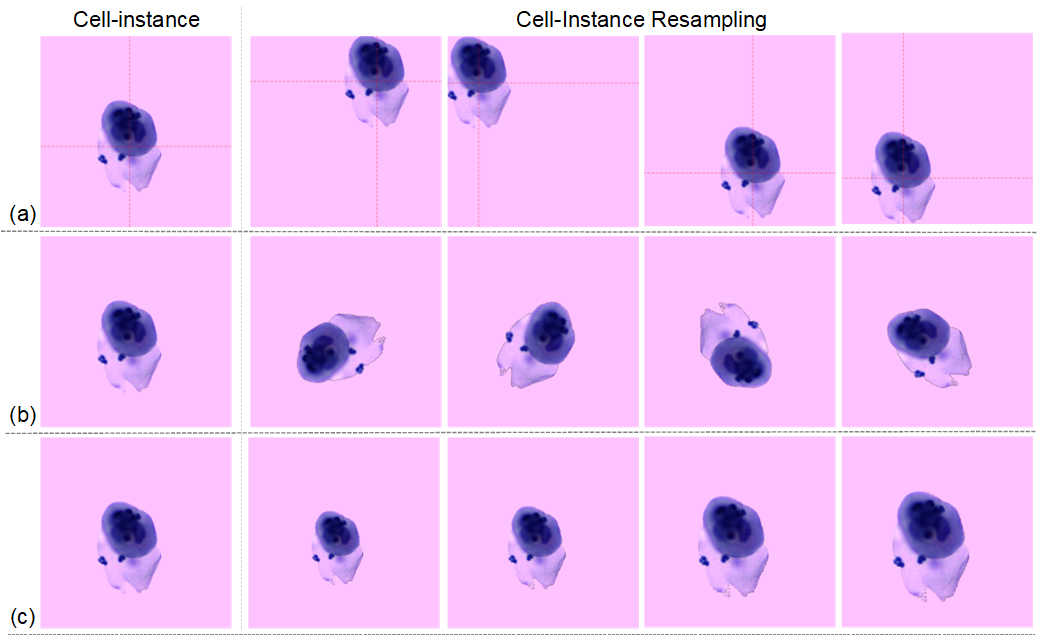}
    \vspace{-3mm}
    \caption{Cell-instance resampling. (a) Random placement of cell-instance; (b) Random cell-instance rotation; (c) Random scaling within a limited range.}
    \label{fig:cell-n}
\end{figure*}

\subsubsection{Cell-Instance Resampling Strategy}

To uniformize the spatial distribution of cell populations, we perform random resampling on all isolated cell instances. We create a blank layout canvas matching the resolution of the original ROI image, with a light-pink background (RGB: 253, 194, 255). This color is statistically summarized from the background tone of original TCT microscopic images, and alternative solid colors can be used flexibly according to practical demands. \textbf{Guided by a unified processing pipeline, all individual cell instances are repositioned and rearranged on the constructed canvas (Illustrations are as shown in Fig.~\ref{fig:cell-n}).}

Each cell instance undergoes geometric augmentation prior to layout arrangement via two sequential random transformations. First, each cell is randomly scaled with a factor uniformly sampled from 0.9 to 1.1. Bilinear interpolation is adopted in the resizing process to preserve pixel authenticity and prevent structural distortion of cell morphology. Second, random rotation is applied at angles ranging from 0° to 360°. To avoid incomplete cell truncation during rotation, we adaptively expand the canvas boundary. All transformation parameters, including scaling factors, rotation angles and affine matrices, are fully recorded to ensure experimental traceability for subsequent analyses.

A non-overlapping placement mechanism is further designed to eliminate spatial overlap among relocated cell patches. For each individual cell, candidate coordinates are randomly sampled, and axis-aligned bounding box collision detection is adopted to judge spatial overlap with existing deployed cells. Once overlap is detected, new coordinates will be resampled iteratively. To secure a reasonable proportion of abnormal cells in training data, abnormal instances are arranged with priority and granted doubled placement attempts. Normal cells are sequentially filled into the remaining canvas space in descending order of patch area, which balances spatial utilisation and rational overall sample distribution. To achieve efficient data expansion with limited computational overhead, SAM segmentation and cell-instance extraction are executed only once for each original image, while the random resamplingt pipeline is repeated multiple times with independent random seeds. \textbf{It is worth noting that each cell entity in the standardized images maintains a one-to-one correspondence with its original counterpart. Accordingly, detection and recognition results obtained from standardized samples can be accurately mapped back to the original images.}

\begin{figure*}[!t]
    \centering
    \includegraphics[width=0.96\textwidth, keepaspectratio]{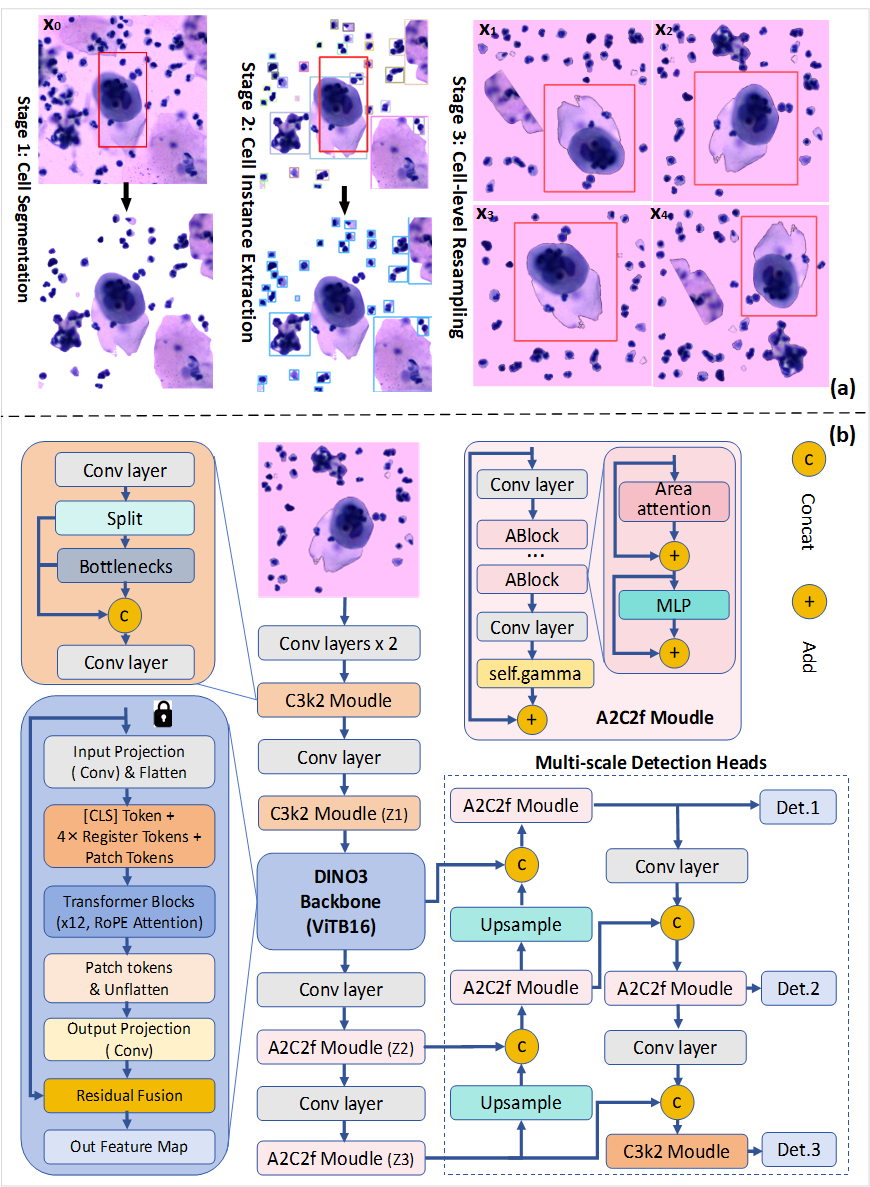}
    \caption{Overview of the proposed method. (a) The pipelines of our Cell-Norm method; (b) The architecture of the proposed YoLo-D model.}
    \label{fig:arch}
\end{figure*}

\subsection{YoLo-D Model}\label{subsec:models}

\subsubsection{Overview of our YoLo-D}
This paper presents a one-stage object detection model that fuses YOLOv12 and DINOv3 models (named YoLo-D, as shown in Fig.~\ref{fig:arch}(b)). The key strength of the proposed model lies in its integration of the powerful global visual representation capabilities of DINOv3 and the advanced detection performance of YOLOv12. Specifically, we incorporate the DINO backbone into the YOLOv12 backbone network, aiming to enhance the model's ability to characterize small-target features in images. The overall framework follows a pipeline of convolutional backbone feature extraction, feature enhancement, multi-scale feature fusion, and detection via a decoupled head.

\subsubsection{Adaptive Design of YOLOv12}
The backbone network is built upon the YOLOv12-Large architecture, with tailored channel adaptation and structural refinement to ensure efficient compatibility with the DINOv3 enhancement module. To strike a favorable balance between detection accuracy and computational efficiency, the maximum number of channels is limited to 512. The network first performs initial downsampling using two standard convolutional layers with stride 2, gradually expanding the channel dimension from 3 to 64 and then to 128, while reducing the feature map size to 1/4 of the input size. Subsequently, the backbone increases the channel count to 256 via two C3k2 blocks. After convolutional downsampling while preserving the channel dimension, a feature map of 1/8 input size is obtained. A second C3k2 block further expands channels to 512, generating the feature maps that are fed into the DINOv3 module. Following DINOv3-based feature enhancement and residual fusion, the backbone maintains 512 channels via a stride-2 convolution and stacks four A2C2f modules to encode high-level features at the Z2 scale. After another downsampling operation, four additional A2C2f modules are stacked to produce the Z3 scale feature maps.

\subsubsection{Adaptive Design of DINOv3}
To enhance the global representation capability of the convolution layer, a pre-trained DINOv3-ViT-B/16 model is inserted after the YOLOv12-Large backbone as the feature enhancement module. This module adopts a weight freezing strategy, serving only as a plug-and-play attention enhancement unit.

First, an input projection layer converts the convolutional feature maps into a sequence format processable by the Transformer: the spatial dimension of the feature maps is flattened into a sequence length, and the number of channels is linearly projected from 512 to 768 to align with the hidden layer dimension of DINOv3-ViT-B/16, resulting in a Patch feature sequence with the shape (batch\_size, 6400, 768).

Subsequently, one learnable CLS Token and four Register Tokens are concatenated at the beginning of the sequence. The CLS Token is used to capture global feature information, while the Register Tokens stabilize the Transformer training process and reduce the interference of noise on feature learning, ultimately forming a complete input sequence containing CLS Token, Register Tokens, and Patch Tokens.

The complete input sequence is fed into the DINOv3-ViT-B/16 backbone, a sequence containing all Tokens is output. The module retains only the Patch Tokens and discards the CLS Token and Register Tokens. The output sequence is restored to a 2D feature map through an Unflatten operation, then mapped back to the convolutional format via an output projection layer---projecting the number of channels from 768 to 512 and reshaping the sequence dimension into the same feature map shape as the input. Finally, residual fusion is performed between this feature map and the original Z1 feature map to achieve feature enhancement and information fusion, and the enhanced Z1 feature map is fed into the subsequent feature fusion network.

\subsubsection{Detection Heads}
The model adopts the native PAN-FPN structure of YOLOv12 as the feature fusion network. Through bidirectional feature transmission paths (top-down and bottom-up), it achieves efficient fusion of features at different scales and enhances the interactive transmission of location information and semantic information in the features. The entire fusion network uses the A2C2f module as the feature refinement unit instead of traditional convolution modules. Finally, the network outputs the enhanced Z1, Z2, and Z3 feature maps at three scales, which are fed into the subsequent detection head for target detection tasks.

Specifically, the Z1 feature maps are responsible for small object detection, Z2 for medium objects, and Z3 for large objects. Each scale of features is separately fed into two parallel branches: the classification branch and the regression branch. The classification branch employs stacked convolutions to predict category probabilities, while the regression branch estimates bounding box coordinates and confidence scores. After obtaining multi-scale prediction results, non-maximum suppression (NMS) is adopted to eliminate redundant bounding boxes, thereby producing the final detection outputs.

\subsubsection{Loss Function}
We employed the YOLO-series object detection loss to train our YoLo-D model. The overall loss consists of three components: bounding box regression loss, classification loss, and class-level distribution focal loss. The total loss function is defined as:

\begin{equation}
\mathcal{L}
=
\lambda_{box}\mathcal{L}_{box}
+
\lambda_{cls}\mathcal{L}_{cls}
+
\lambda_{dfl}\mathcal{L}_{dfl}
\label{eq:loss}
\end{equation}

where $\mathcal{L}_{box}$ denotes the bounding box regression loss, $\mathcal{L}_{cls}$ denotes the classification loss, and $\mathcal{L}_{dfl}$ denotes the distribution focal loss. $\lambda_{box}$, $\lambda_{cls}$, and $\lambda_{dfl}$ are the corresponding weighting coefficients. In this study, these three weights are set to 7.5, 0.5, and 1.5, respectively.

\textbf{(1) Bounding Box Regression Loss} is used to measure the differences in position, scale, and shape between the predicted and the ground-truth bounding box. In this study, the CIoU loss~\cite{zheng2020distance} is adopted as the bounding box regression loss:

\begin{equation}
\mathcal{L}_{box}
=
1-\text{CIoU}(B,\hat{B})
\label{eq:ciou_loss}
\end{equation}

where $B$ denotes the ground-truth bounding box and $\hat{B}$ denotes the predicted bounding box. The CIoU is calculated as:

\begin{equation}
\text{CIoU}
=
\text{IoU}
-
\frac{\rho^2(\mathbf{b},\hat{\mathbf{b}})}{c^2}
-
\alpha v
\label{eq:ciou}
\end{equation}

where $\text{IoU}$ is the intersection over union between the predicted ($\hat{B}$) and ground-truth boxes ($B$); $\rho^2(\mathbf{b},\hat{\mathbf{b}})$ denotes the squared Euclidean distance between the center points; $c$ is the diagonal length of the smallest enclosing box; $v$ measures the consistency of the aspect ratio; and $\alpha$ is a balancing coefficient.

\begin{equation}
v
=
\frac{4}{\pi^2}
\left(
\arctan \frac{w}{h}
-
\arctan \frac{\hat{w}}{\hat{h}}
\right)^2
\label{eq:v}
\end{equation}

\begin{equation}
\alpha
=
\frac{v}{1-\text{IoU}+v}
\label{eq:alpha}
\end{equation}

\textbf{(2) Classification Loss} is used to measure the discrepancy between the predicted class and the ground-truth labels. Binary cross-entropy loss is adopted as the classification loss:

\begin{equation}
\mathcal{L}_{cls}
=
-\sum_{c=1}^{C}
\left[
y_c\log(\hat{p}_c)
+
(1-y_c)\log(1-\hat{p}_c)
\right]
\label{eq:cls}
\end{equation}

where $C$ denotes the total number of classes, $y_c$ is the ground-truth label of class $c$, and $\hat{p}_c$ is the predicted probability that the sample belongs to class $c$.

\textbf{(3) Distribution Focal Loss}~\cite{li2022generalized} is used to improve bounding box localization accuracy. YOLOv12 models the regression values of the four bounding box sides as discrete probability distributions rather than directly regressing continuous coordinates. Let $y$ denote the continuous ground-truth distance. Its left and right adjacent discrete positions are defined as:

\begin{equation}
y_l=\lfloor y \rfloor,\quad y_r=y_l+1
\label{eq:ylr}
\end{equation}

The distribution focal loss is then defined as:

\begin{equation}
\mathcal{L}_{dfl}
=
-(y_r-y)\log(P(y_l))
-
(y-y_l)\log(P(y_r))
\label{eq:dfl}
\end{equation}

where $P(y_l)$ and $P(y_r)$ denote the predicted probabilities that the distance falls into the left and right discrete positions, respectively. By applying weighted constraints to the two adjacent discrete positions, DFL enables a more precise representation of bounding box boundaries and thus improves localization accuracy.

\section{Experiments}
\subsection{Data Preparation}
All experiments in this study were conducted on a public TCT dataset, which contains 8037 original 2048$\times$2048 resolution TCT images and comprises a total of 14684 annotated abnormal cells. For subsequent multi-scale evaluation, we cropped local ROI patches of varying sizes, including 640$\times$640, 1024$\times$1024, and 1536$\times$1536, around on each annotated abnormal cell, yielding 14684 cell-containing image patches in total. These cropped ROI samples were randomly split into training, validation, and test sets at a ratio of 7:1:2 to construct the original baseline TCT dataset. Based on this baseline data, we further applied the proposed Cell-Norm method to implement cell-level data augmentation. The resulting standardized dataset was adopted for model training in all subsequent experimental evaluations.

\subsection{Training Configurations}
Our proposed model is constructed based on YOLOv12-L, integrated with a DINOv3-ViT-B/16 visual encoder. The DINOv3 encoder is initialized with publicly available self-supervised pre-trained DINOv3 weights and fixed throughout training; learnable parameters are restricted to the YOLOv12 backbone, neck, detection head, as well as the cross-modal fusion module bridging DINOv3 and the YOLO detection pipeline. We adopt SGD as the optimization solver with an initial learning rate of 0.01 and a final learning rate of 0.0001 ($\textit{lrf} = 0.01$), alongside a momentum of 0.937 and weight decay set to $5 \times 10^{-4}$. Learning rate decays linearly after three warm-up epochs, where warm-up momentum and warm-up bias learning rate are configured as 0.8 and 0.1 respectively. The nominal batch size for loss scaling ($\textit{nbs}$) is fixed at 64, and all input images are resized to 640$\times$640 pixels. The network is trained over 100 epochs with an actual batch size of 32 (16 samples per single GPU). Default data augmentation rules from the original YOLOv12 implementation are strictly followed. Mixed-precision training (AMP) is turned off and full FP32 precision is enforced for the whole training procedure to avoid numerical instability during DINOv3 feature extraction. All training experiments run on two NVIDIA A100 GPUs via distributed data parallelism (DDP), with 16 dataloader workers and a fixed random seed of 0 for reproducibility.

\subsection{Results Analysis}

\subsubsection{Comparisons with the Classic Methods}

\begin{figure}[htbp]
    \centering
    \includegraphics[width=\textwidth]{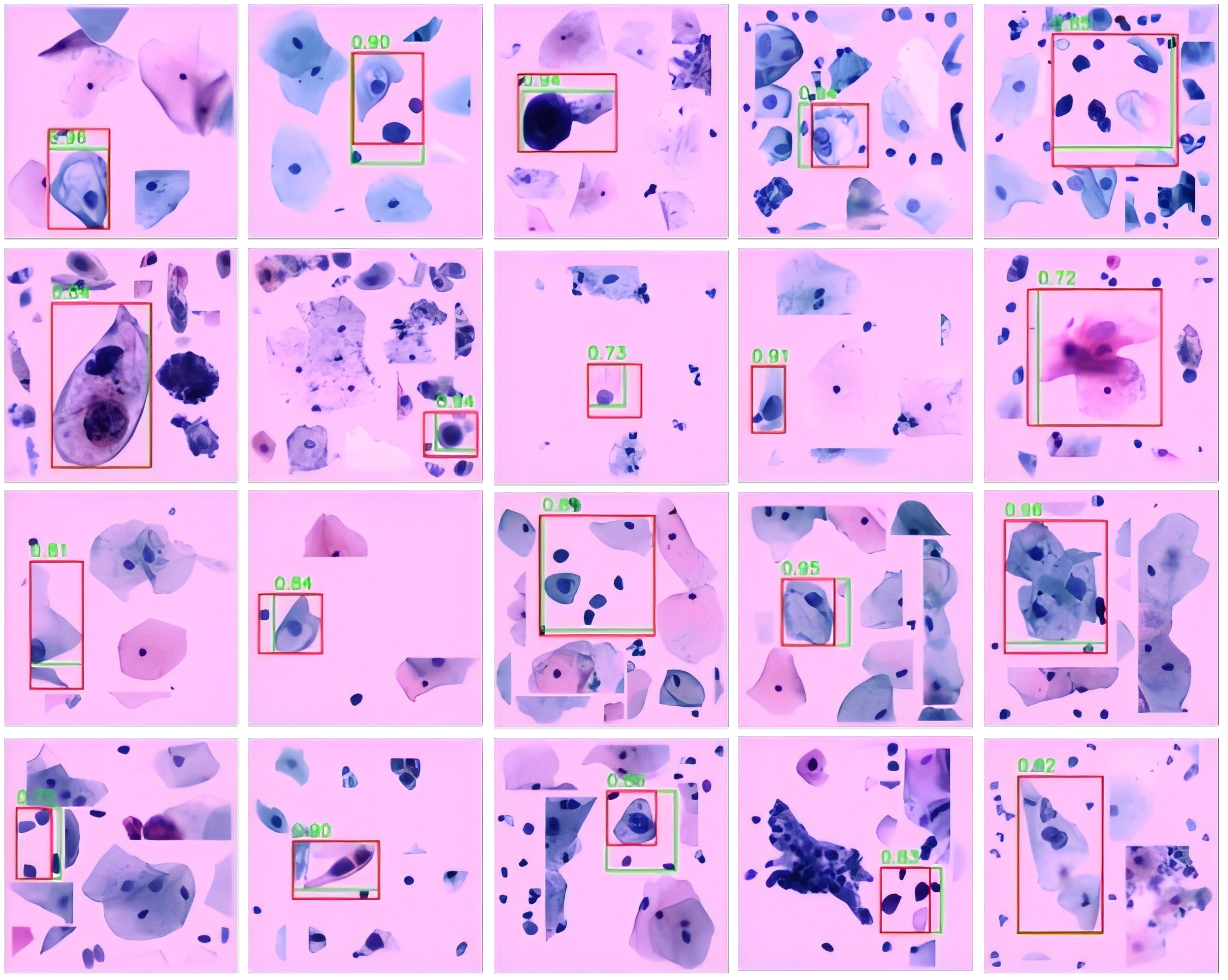}
    \vspace{-3mm}
    \caption{Illustrations of TCT image detection and recognition based on the proposed YoLo-D model and cell-norm method.}
    \label{fig:res1}
\end{figure}

Tab.~\ref{tab:sota} quantitatively compares the performance of various methods (SSD~\cite{liu2016ssd}, RetinaNet, FCOS~\cite{tian2019fcos}, Faster R-CNN, Cascade R-CNN~\cite{cai2018cascade}, Sparse R-CNN~\cite{sun2021sparse}, YOLOv3~\cite{redmon2018yolov3}, YOLOv7~\cite{wang2023yolov7}, DETR~\cite{carion2020end}, and YOLOv12) on the TCT image detection and recognition task (some illustrations are demonstrated in Fig.~\ref{fig:res1}). Results demonstrate that the proposed method achieves the state-of-the-art performance across all measurement metrics, and evidently outperforms other existing approaches by a considerable margin.

\begin{table}[htbp]
\caption{Comparisons with various methods. The proposed YOLO-D model with 4$\times$ data expansion using our Cell-Norm method achieves the best performance across all evaluation metrics (All models were trained using 640$\times$640 ROIs cropped from the original 2048$\times$2048 TCT images). Best results are highlighted in bold.}\label{tab:sota}%
\begin{tabular*}{\textwidth}{@{\extracolsep\fill}lccccc@{\extracolsep\fill}}
\toprule
Method & $AP_{50:95}$ & $AP_{50}$ & $AP_{75}$ & $AR_{50:95}$ & F1-score \\
\midrule
SSD~\cite{liu2016ssd}                & 10.8  & 24.1  & 7.2   & 14.3  & 40.1  \\
RetinaNet~\cite{lin2017focal}       & 25.7  & 54.8  & 20.3  & 34.4  & 66.5  \\
FCOS~\cite{tian2019fcos}            & 27.6  & 61.6  & 21.2  & 37.7  & 68.9  \\
Faster R-CNN~\cite{ren2015faster}   & 31.5  & 67.4  & 25.0  & 45.2  & 58.9  \\
Cascade R-CNN~\cite{cai2018cascade} & 29.1  & 57.7  & 27.0  & 39.2  & 66.4  \\
Sparse R-CNN~\cite{sun2021sparse}   & 23.2  & 50.1  & 19.0  & 32.0  & 65.8  \\
YOLOv3~\cite{redmon2018yolov3}      & 9.1   & 28.3  & 2.8   & 17.6  & 46.3  \\
YOLOv7~\cite{wang2023yolov7}        & 13.3  & 37.3  & 5.4   & 22.6  & 55.3  \\
DETR~\cite{carion2020end}           & 19.4  & 47.1  & 12.2  & 36.5  & 43.7  \\
YOLOv12~\cite{tian2026yolov12}      & 48.09 & 83.54 & 50.62 & 81.78 & 77.01 \\
YOLO-D (1$\times$) (Ours)           & 85.83 & 87.90 & 86.39 & 98.36 & 79.79 \\
\textbf{YOLO-D (4$\times$) (Ours)}   & \textbf{92.41} & \textbf{94.32} & \textbf{92.67} & \textbf{98.76} & \textbf{86.87} \\
\bottomrule
\end{tabular*}
\end{table}

From the perspective of task properties and existing technical limitations, most current studies on biomedical cell detection and recognition primarily focus on optimizing the network architectures and loss functions of generic detection frameworks, while largely overlooking the inherent characteristics of medical cell images. In TCT cytological imaging scenarios, cell populations exhibit substantial variations in density, morphology, scale, and aggregation degree. Such inherent discrepancies prevent deep learning models from learning stable and robust feature representations, ultimately leading to degraded generalization performance. Rather than merely enhancing model feature representation capability, mitigating the intrinsic distribution variations of cell instances in medical images is more fundamental and practically significant for improving clinical detection performance. To address this issue, the proposed Cell-Norm method standardizes cell-instance distribution across both training and test TCT images, achieving consistent independent and identically distributed data characteristics. Benefiting from this optimized data distribution, our YOLO-D model achieves superior performance over competing methods with clear margins, which fully validates the effectiveness and robustness of our approach for medical cell detection and recognition tasks.

\begin{figure}[h]
    \centering
    \includegraphics[width=\textwidth]{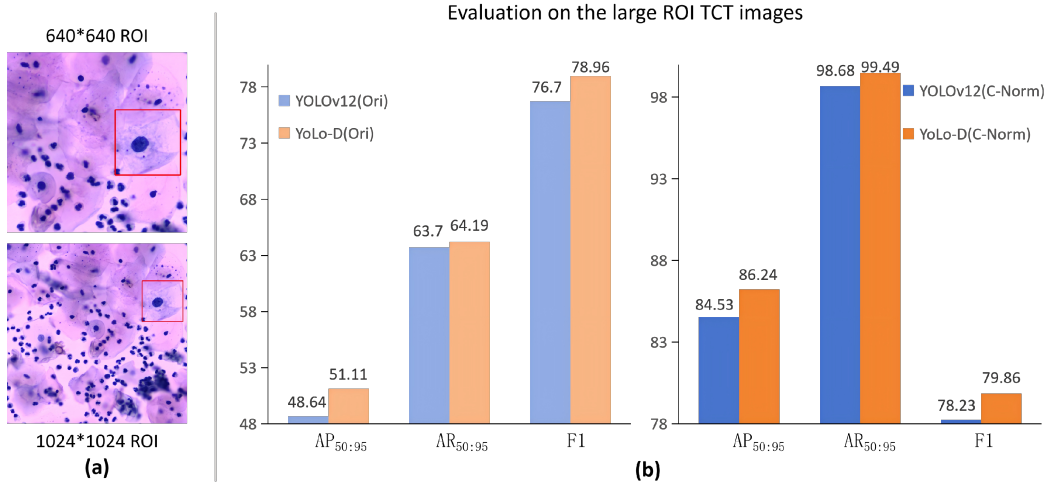}
    \vspace{-3mm}
    \caption{Comparisons on different scale ROI TCT images. (a) Illustrations of 1024$\times$1024 and 640$\times$640 TCT images; (b) Performance comparison.}
    \label{fig:Comparisons}
\end{figure}

\subsubsection{Evaluation on Large ROI TCT Images}
We comprehensively evaluated the performance of our proposed YOLO-D model and the baseline YOLOv12 on 1024$\times$1024 ROI patches cropped from raw TCT images. Compared with 640$\times$640 ROI samples, larger patches accommodate more cellular regions and complex background information, substantially increasing the difficulty of abnormal cell recognition, as visualized in Fig.~\ref{fig:Comparisons}(a). From the results in Table~\ref{tab:generalization} and Fig.~\ref{fig:Comparisons}(b), we have made three critical findings. First, both models exhibit a performance drop on large-scale ROI images, confirming the higher complexity of such challenging detection scenarios. Second, performance of both models improves remarkably after adopting our Cell-Norm method on original TCT data, which verifies the efficacy of the proposed method. Third, our YOLO-D model achieves more competitive performance than the advanced YOLOv12 on difficult detection tasks, demonstrating that the embedded DINO module can effectively strengthen the model's discriminative feature representation ability.

\begin{table}[htbp]
    \caption{Comparisons of training YoLo-D and YOLOv12 models with 1024$\times$1024 ROI image data. ``Ori.'' denotes the models trained on the original TCT image; ``C-Norm'' denotes the models trained on the normalization TCT images.}\label{tab:generalization}%
    \begin{tabular*}{\textwidth}{@{\extracolsep\fill}lccccc@{\extracolsep\fill}}
    \toprule
    Method & $AP_{50:95}$ & $AP_{50}$ & $AP_{75}$ & $AR_{50:95}$ & F1-score \\
    \midrule
    YOLOv12 (Ori.)             & 48.64 & 83.08 & 52.43 & 63.70 & 76.70 \\
    YoLo-D (Ori.)              & 51.11 & 85.92 & 55.97 & 64.19 & 78.96 \\
    \textbf{YOLOv12 (C-Norm)}  & \textbf{84.53} & \textbf{85.64} & \textbf{84.49} & \textbf{98.68} & \textbf{78.23} \\
    \textbf{YoLo-D (C-Norm)}   & \textbf{86.24} & \textbf{87.10} & \textbf{86.22} & \textbf{99.49} & \textbf{79.86} \\
    \bottomrule
    \end{tabular*}
\end{table}

\subsubsection{Robustness Evaluation}

\begin{figure}[htbp]
\centering
\includegraphics[width=\textwidth]{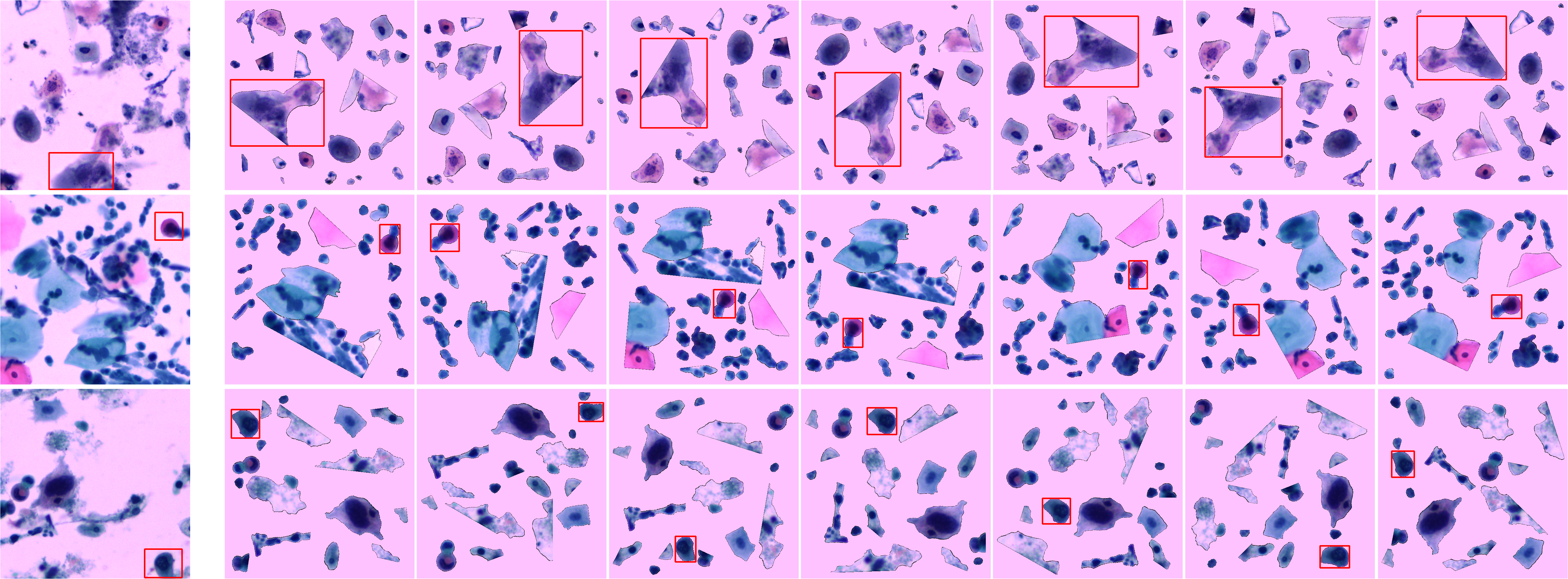}
\vspace{-3mm}
\caption{Visual demonstration between original TCT images and their corresponding seven-times C-Norm augmented images. Model robustness is further evaluated on these multi-version C-Norm-processed TCT images.}
\label{fig:Visual_comparison}
\end{figure}

The proposed Cell-Norm method relies on the random resampling of individual cell instances to regularize cell-population distribution within the TCT images. As a result, it inevitably leads to variations in the spatial arrangement of cell instances among synthetic samples generated in different trials. To quantify how such random fluctuations affect detection accuracy, we produce seven augmented variants per raw test image and build seven independent test subsets for this robustness evaluation (some intuitive visual illustrations are shown in Fig.~\ref{fig:Visual_comparison}). The results in Table~\ref{tab:robustness} reveal that random resampling brings only slight fluctuations to final detection metrics. These limited performance shifts can be safely ignored under practical experimental settings, which verifies the proposed scheme is stable and dependable for abnormal cell recognition and detection in TCT images.

\begin{table}[htbp]
    \caption{Robustness evaluation of the proposed YOLO-D model on seven independent augmented test sets. The last row presents the overall mean $\pm$ standard deviation calculated across all seven evaluation runs.}\label{tab:robustness}%
    \footnotesize
    \begin{tabular*}{\textwidth}{@{\extracolsep\fill}lccccc@{\extracolsep\fill}}
    \toprule
    Method & $AP_{50:95}$ & $AP_{50}$ & $AP_{75}$ & $AR_{50:95}$ & F1-score \\
    \midrule
    YoLo-D (test0) & 92.24 & 94.07 & 92.96 & 98.45 & 86.87 \\
    YoLo-D (test1) & 92.69 & 94.50 & 93.00 & 98.68 & 87.64 \\
    YoLo-D (test2) & 92.40 & 94.22 & 92.64 & 98.37 & 87.59 \\
    YoLo-D (test3) & 91.97 & 93.89 & 92.36 & 98.40 & 86.83 \\
    YoLo-D (test4) & 92.80 & 94.88 & 93.21 & 98.55 & 88.03 \\
    YoLo-D (test5) & 92.44 & 94.24 & 93.06 & 98.47 & 87.83 \\
    YoLo-D (test6) & 91.85 & 93.72 & 92.29 & 98.43 & 87.08 \\
    \midrule
    \textbf{Mean $\pm$ Std}
        & $92.34 \pm 0.35$
        & $94.22 \pm 0.39$
        & $92.79 \pm 0.36$
        & $98.48 \pm 0.11$
        & $87.41 \pm 0.48$ \\
    \bottomrule
    \end{tabular*}
\end{table}

\subsection{Ablation Study}
This study delivers two major contributions to TCT image detection. First, unlike conventional image-level data augmentation method that merely manipulate pixel information, we propose a cell populations distribution normalization method to standardize the spatial distribution of cell instances within TCT images, addressing the inherent data heterogeneity of cytological samples. Second, we integrate the cutting-edge object detection framework YOLOv12 with the powerful general visual representation backbone DINOv3 to build a robust cell detection network. Comprehensive ablation experiments are further conducted to quantitatively validate the effectiveness and rationality of the above-proposed designs.

\begin{figure}[htbp]
    \centering
    \includegraphics[width=\textwidth]{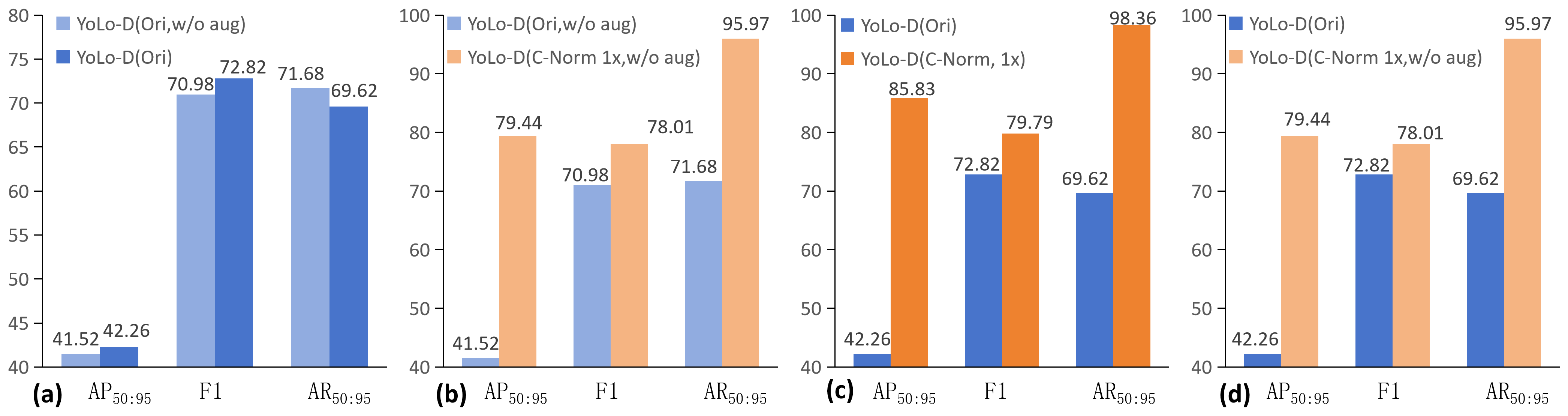}
    \vspace{-3mm}
    \caption{Comparisons of conventional image augmentation and the proposed C-Norm method.}
    \label{fig:zhuxingtu1}
\end{figure}

(1) In Tab.~\ref{tab:aug_ablation}, ``Ori'' denotes the YOLO-D model trained on the original TCT dataset, while ``w/o aug'' represents model training implemented without conventional data augmentation techniques, including Mosaic, MixUp, and HSV color jitter. ``C-Norm'' refers to the YOLO-D model trained on the TCT dataset standardized via our cell normalization strategy. Experimental results demonstrate that traditional image-level augmentation only yields marginal performance improvements for the abnormal cell detection and classification tasks (Group 1 vs. Group 2, Fig.~\ref{fig:zhuxingtu1}(a)). In contrast, the proposed C-Norm method achieves substantial performance gains across all evaluation metrics compared with baseline schemes (Group 1 vs. Group 3, Fig.~\ref{fig:zhuxingtu1}(b); Group 2 vs. Group 4, Fig.~\ref{fig:zhuxingtu1}(c)). Notably, significant performance promotion can be exclusively attributed to the C-Norm strategy, even when all conventional augmentation operations are disabled (Group 2 vs. Group 3, Fig.~\ref{fig:zhuxingtu1}(d)).

\begin{table}[htbp]
\caption{Ablation experiments for YoLo-D models were carried out to confirm the performance gains brought by conventional image augmentation and the proposed C-Norm method.}\label{tab:aug_ablation}%
\small
\begin{tabular*}{\textwidth}{@{\extracolsep\fill}lccccc@{\extracolsep\fill}}
\toprule
Method & $AP_{50:95}$ & $AP_{50}$ & $AP_{75}$ & $AR_{50:95}$ & F1-score \\
\midrule
1: YoLo-D (Ori, w/o aug)        & 41.52 & 74.47 & 42.31 & 71.68 & 70.98 \\
2: YoLo-D (Ori)                 & 42.26 & 72.79 & 45.57 & 69.62 & 72.82 \\
3: YoLo-D (C-Norm 1$\times$, w/o aug) & 79.44 & 83.17 & 80.26 & 95.97 & 78.01 \\
4: YoLo-D (C-Norm, 1$\times$)   & 85.83 & 87.90 & 86.39 & 98.36 & 79.79 \\
\bottomrule
\end{tabular*}
\end{table}

\begin{figure}[htbp]
    \centering
    \includegraphics[width=\textwidth]{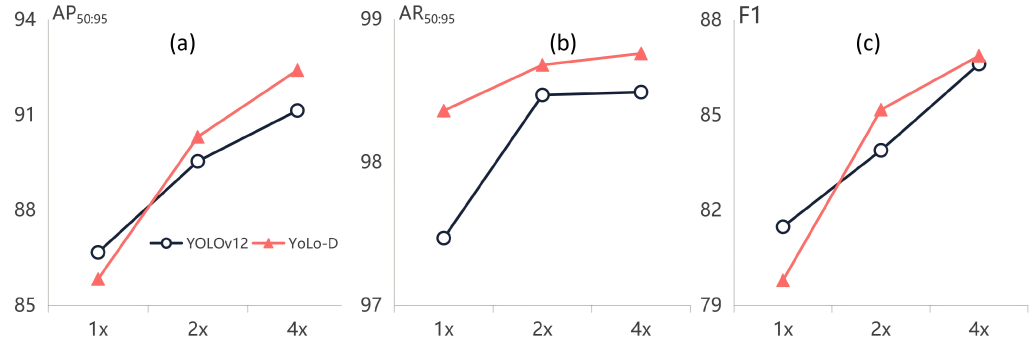}
    \vspace{-3mm}
    \caption{Comparisons of the proposed YoLo-D and YOLOv12 models with training data expansion via the proposed Cell-Norm method.}
    \label{fig:zhexiantu}
\end{figure}

(2) Gradual expansion of the original dataset at scaling ratios of 1$\times$, 2$\times$, and 4$\times$ via the proposed Cell-Norm method yields consistent and steady performance improvements across all evaluation metrics (See Tab.~\ref{tab:yolo-d_124} and Fig.~\ref{fig:zhexiantu}). This phenomenon stems from the inherent drawbacks of conventional data augmentation strategies. Traditional augmentation approaches merely perform pixel-level or global image transformation and fail to optimize the spatial layout of cell populations in TCT images. In comparison, our method targets the optimization of cell spatial distribution, achieving uniform and consistent cell arrangement in both training and test samples. This optimization strictly satisfies the fundamental prerequisite for reliable model training, i.e., the independent and identically distributed (i.i.d.) assumption between training and test datasets. Overall, the proposed Cell-Norm strategy maximizes compliance with this core statistical assumption, thereby effectively facilitating stable and robust model learning for TCT image detection tasks.

\begin{table}[htbp]
    \caption{Performance of the YoLo-D model with training data expansion via the Cell-Norm method.}%
    \begin{tabular*}{\textwidth}{@{\extracolsep\fill}lccccc@{\extracolsep\fill}}
    \toprule
    Method & $AP_{50:95}$ & $AP_{50}$ & $AP_{75}$ & $AR_{50:95}$ & F1-score \\
    \midrule
    YoLo-D (1$\times$) & 85.83 & 87.90 & 86.39 & 98.36 & 79.79 \\
    YoLo-D (2$\times$) & 90.31 & 92.46 & 90.39 & 98.68 & 85.17 \\
    \textbf{YoLo-D (4$\times$)}
        & \textbf{92.41} & \textbf{94.32} & \textbf{92.67}
        & \textbf{98.76} & \textbf{86.87} \\
    \bottomrule
    \end{tabular*}
    \label{tab:yolo-d_124}
\end{table}

(3) We further validate the compatibility of our Cell-Norm method by applying it to the advanced YOLOv12 baseline model. Consistent results (See Tab.~\ref{tab:yolov12_124} and Fig.~\ref{fig:zhexiantu}) reveal that the proposed method can substantially improve the performance of YOLOv12 model. This cross-model performance improvement verifies the excellent generalization and robustness of our method, demonstrating its reliable applicability to mainstream object detection architectures.

\begin{table}[htbp]
\caption{Performance analysis of the proposed YOLOv12 model with training data expansion via the proposed Cell-Norm method.}%
\begin{tabular*}{\textwidth}{@{\extracolsep\fill}lccccc@{\extracolsep\fill}}
\toprule
Method & $AP_{50:95}$ & $AP_{50}$ & $AP_{75}$ & $AR_{50:95}$ & F1-score \\
\midrule
YOLOv12 (1$\times$) & 86.67 & 89.49 & 87.50 & 97.47 & 81.48 \\
YOLOv12 (2$\times$) & 89.54 & 92.07 & 89.87 & 98.47 & 83.89 \\
\textbf{YOLOv12 (4$\times$)}
    & \textbf{91.14} & \textbf{93.44} & \textbf{91.60}
    & \textbf{98.49} & \textbf{86.61} \\
\bottomrule
\end{tabular*}
\label{tab:yolov12_124}
\end{table}

\begin{figure}[htbp]
    \centering
    \includegraphics[width=\textwidth]{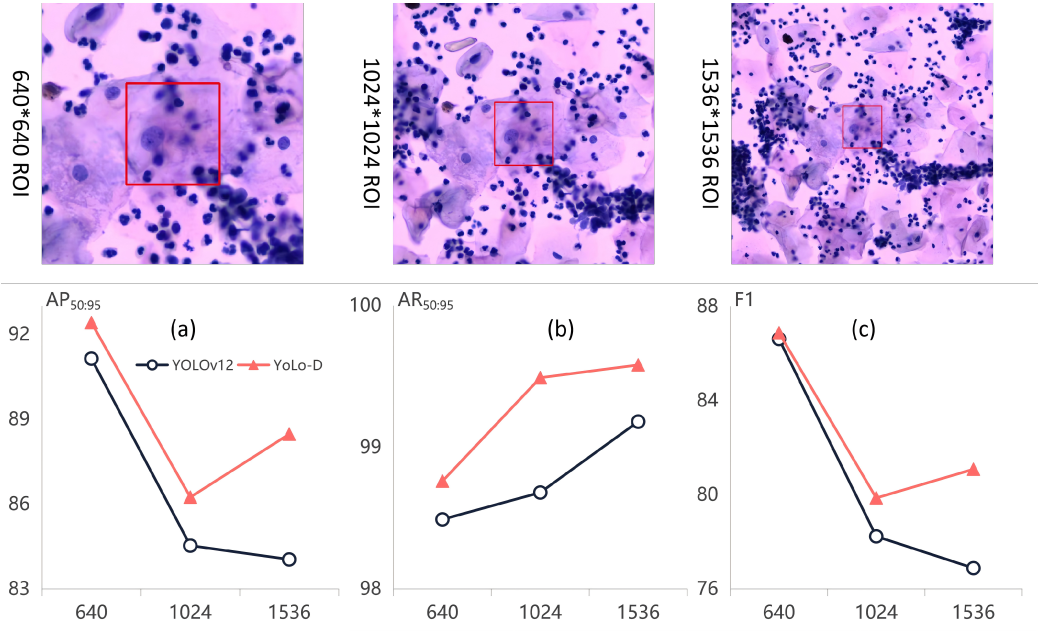}
    \vspace{-3mm}
    \caption{Visual comparisons of the proposed YoLo-D and YOLOv12 models on cross-scale ROI images.}
    \label{fig:pic_table7}
\end{figure}

(4) We benchmark the proposed YOLO-D and baseline YOLOv12 models on TCT image task with multi-scale ROI crops of 640$\times$640, 1024$\times$1024, and 1536$\times$1536 (See Fig.~\ref{fig:pic_table7}), where larger ROI images correspond to more challenging detection and recognition tasks. From the results in Table~\ref{tab:full_compare} and Fig.~\ref{fig:pic_table7}, we have made several key conclusions: First, our YoLo-D model along with the cell-norm method exhibits only marginal performance advantages with limited training data (1$\times$). Nevertheless, it achieves substantial performance gains and thoroughly outperforms the state-of-the-art YOLOv12 as the training data scale expands to 2$\times$ and 4$\times$. Second, the proposed YOLO-D model yields significantly better performance than YOLOv12 on high-difficulty tasks with large-scale TCT ROIs (both 1024$\times$1024 and 1536$\times$1536), an effect particularly beneficial for the challenging task. Furthermore, experimental results indicate that the performance improvement delivered by our cell-instance resampling strategy far exceeds the benefits brought by simple network structural optimization.

\begin{table}[htbp]
    \caption{Comparisons of the proposed YoLo-D model and YOLOv12 models. Both the models employ the proposed cell-norm method.}\label{tab:full_compare}%
    \small
    \begin{tabular*}{\textwidth}{@{\extracolsep\fill}clccccc@{\extracolsep\fill}}
    \toprule
    ROI Size & Method & $AP_{50:95}$ & $AP_{50}$ & $AP_{75}$ & $AR_{50:95}$ & F1-score \\
    \midrule
    \multirow{6}{*}{640$\times$640}
        & YOLOv12 (1$\times$)        & 86.67 & 89.49 & 87.50 & 97.47 & 81.48 \\
        & YoLo-D (1$\times$)         & 85.83 & 87.90 & 86.39 & 98.36 & 79.79 \\
    \cdashline{2-7}
    \noalign{\vskip 3pt}
        & YOLOv12 (2$\times$)        & 89.54 & 92.07 & 89.87 & 98.47 & 83.89 \\
        & YoLo-D (2$\times$)         & 90.31 & 92.46 & 90.39 & 98.68 & 85.17 \\
    \cdashline{2-7}
    \noalign{\vskip 3pt}
        & YOLOv12 (4$\times$)        & 91.14 & 93.44 & 91.60 & 98.49 & 86.61 \\
        & \textbf{YoLo-D (4$\times$)}& \textbf{92.41} & \textbf{94.32} & \textbf{92.67} & \textbf{98.76} & \textbf{86.87} \\
    \midrule
    \multirow{2}{*}{1024$\times$1024}
        & YOLOv12 (1$\times$)        & 84.53 & 85.64 & 84.49 & 98.68 & 78.23 \\
        & YoLo-D (1$\times$)         & 86.24 & 87.10 & 86.22 & 99.49 & 79.86 \\
    \midrule
    \multirow{2}{*}{1536$\times$1536}
        & YOLOv12 (1$\times$)        & 84.04 & 84.48 & 84.10 & 99.18 & 76.88 \\
        & YoLo-D (1$\times$)         & 88.47 & 88.68 & 88.46 & 99.58 & 81.08 \\
    \bottomrule
    \end{tabular*}
\end{table}

\subsection{Comparisons with Clinical Doctors}

We conducted comparative experiments to evaluate the proposed Cell-Norm method and YOLO-D model against experienced clinicians on TCT abnormal cell recognition task. We randomly selected some TCT images (187 samples) from the official test set, each containing exactly one abnormal cell. Three senior attending physicians who specialize in gynecological cytopathology were invited to participate in blind assessment without access to any additional auxiliary information. In this trial, we only judged whether abnormal cells were correctly identified, while ignoring discrepancies in manually annotated bounding boxes. By comparing manual diagnoses with model predictions, we validated the competitiveness and practical clinical value of our AI approach for routine cervical screening. As illustrated in Fig.~\ref{fig:matrix_total}, we have made two major conclusions:

\begin{figure}[htbp]
    \centering
    \includegraphics[width=\textwidth]{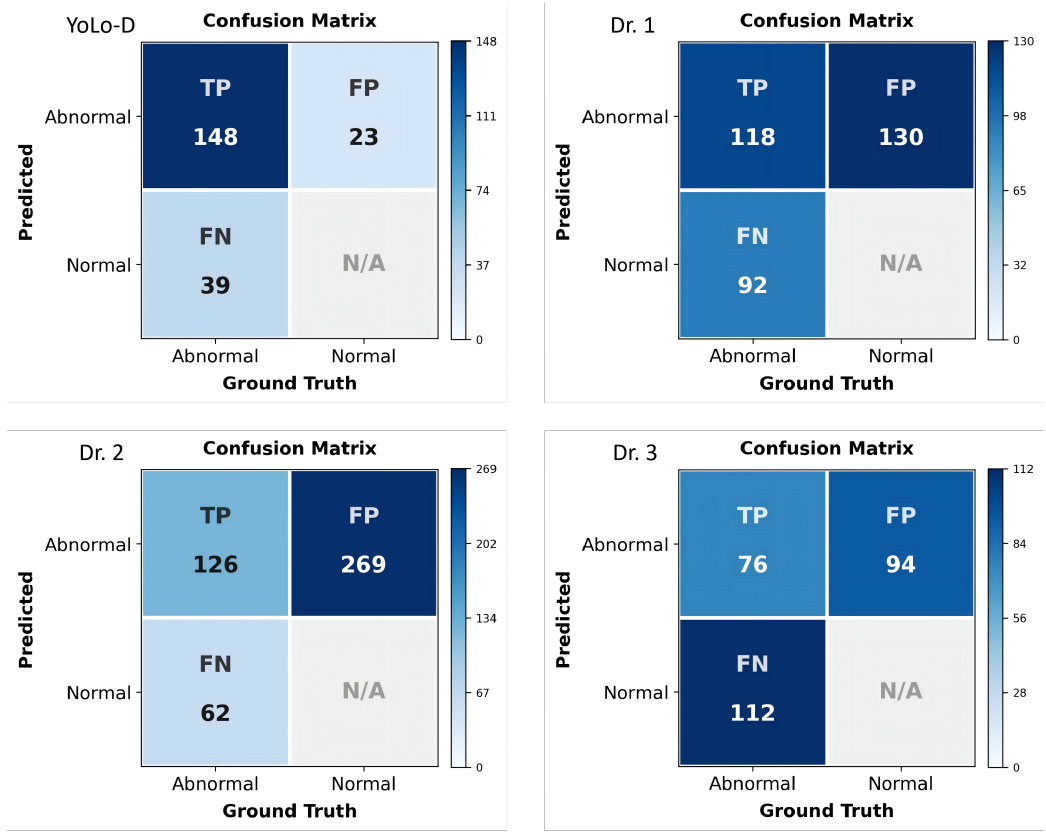}
    \vspace{-3mm}
    \caption{Confusion matrices of the proposed YoLo-D model and three pathologists (Dr.1, Dr.2, Dr.3) on cervical abnormal cell classification task.}
    \label{fig:matrix_total}
\end{figure}

(1) Our method achieves superior overall recognition performance. It yields far more true positives (TP), as well as markedly fewer false positives (FP) and false negatives (FN), compared with clinician assessments. (2) In addition, the counts of TP, FP and FN vary significantly across individual clinicians. This demonstrates that subjective judgment leads to obvious discrepancies in manual diagnosis.

This experiment does not intend to demonstrate that our model can replace clinical physicians. In practice, the interpretation of TCT images is susceptible to clinicians' subjective judgment. There are considerable discrepancies in recognition results among different physicians, and even the same clinician may produce inconsistent diagnoses for the same image at different times. Our model is trained on datasets annotated by clinical physicians, yet we have no way to verify whether those annotators are more authoritative than the clinicians participating in this experiment. For this reason, we cannot conclude that its predictions are inherently superior to manual diagnoses. Nevertheless, if the model is trained on TCT images with authoritative and standardized annotations, its detection results will be more reliable than manual assessment.

\section{Discussion}

This study addresses a critical yet under-explored issue in AI-assisted medical cell analysis: the inconsistent spatial distribution of cell-population in raw images is the primary cause of poor model generalization. Unlike common viewpoints that attribute performance limitations solely to insufficient dataset scale or annotation quality or models, our study confirms that irregular cell distribution across training and testing data violates the independent and identically distributed assumption, fundamentally restricting the practical performance of mainstream vision models. Traditional image-level augmentation methods cannot optimize the internal spatial arrangement of cell instances, which explains why such methods deliver limited performance gains in cytological analysis tasks.

To address this core limitation, we propose a Cell-Norm method to standardize cell-instance distribution in TCT images. Combined with the efficient YOLOv12 detection framework and the powerful feature representation capability of the DINOv3 module, comprehensive results validate that our method achieves state-of-the-art performance, outperforming conventional augmentation strategies and existing baseline models. Further clinical validation demonstrates that the proposed model achieves diagnostic accuracy comparable to experienced pathologists, verifying its effectiveness and clinical practicability for auxiliary cervical cancer screening.

Importantly, the core innovation of our cell-instance normalization is not limited to TCT cervical cytology tasks. Uneven and inconsistent cellular spatial distribution is a pervasive problem across most medical cell imaging datasets. As our method optimizes data quality at the cell-level rather than performing superficial image transformation, it can standardize data distribution for various cytological analysis scenarios. This provides a universal and effective solution for resolving the generalization bottleneck of medical vision models. Therefore, the proposed strategy holds great potential as a general data augmentation paradigm for diverse medical cell analysis tasks.

\section{Acknowledgements}

This work was supported by Scientific and Technological Research Program of Chongqing Municipal Education Commission (No.KJQN202600117), and the Chongqing Science and Technology Bureau, Key Project of the Intelligent Medical Equipment and Frontier Technology Program (No.CSTB2025TIAD-KPX0006).

\bibliographystyle{elsarticle-num}
\bibliography{reference}

\end{document}